\documentclass[letterpaper, 10 pt, conference]{ieeeconf}  %

\IEEEoverridecommandlockouts                              %

\usepackage{graphics} %
\usepackage{epsfig} %
\usepackage{mathptmx} %
\usepackage{times} %
\usepackage{amsmath} %
\usepackage{amssymb}  %
\usepackage{bm}
\usepackage{eucal}
\usepackage{cite}
\usepackage{pgfplots}
\usepackage[T1]{fontenc}
\pgfplotsset{compat=newest}
\usepgfplotslibrary{statistics, fillbetween} %
\usepackage{algorithmic}
\usepackage{tikz}
\usetikzlibrary{arrows.meta, positioning, calc, decorations.pathmorphing, patterns, spy, pgfplots.groupplots, math, intersections, external}
\usepackage{tikz-3dplot}
\usepackage{textcomp}
\usepackage{xcolor}
\usepackage{siunitx}
\usepackage{todonotes}
\usepackage{booktabs}
\usepackage{multirow}
\usepackage{threeparttable}
\usepackage[breaklinks=true, hyperfootnotes=true]{hyperref}
\newcommand\copyrighttext{%
    \footnotesize \textcopyright © 2026 IEEE.  Personal use of this material is permitted.  Permission from IEEE must be obtained for all other uses, in any current or future media, including reprinting/republishing this material for advertising or promotional purposes, creating new collective works, for resale or redistribution to servers or lists, or reuse of any copyrighted component of this work in other works.
}
\newcommand\copyrightnotice{%
    \tikzset{external/export=false}
    \begin{tikzpicture}[remember picture,overlay]
    \node[anchor=south,yshift=10pt, xshift=10pt] at (current page.south) {\fbox{\parbox{\dimexpr\textwidth-\fboxsep-\fboxrule\relax}{\copyrighttext}}};
    \end{tikzpicture}%
    \tikzset{external/export=true}
}
\definecolor{TUMBlue}{rgb}{0.0, 0.40, 0.74}
\definecolor{TUMGray1}{rgb}{0.2, 0.2, 0.2}
\definecolor{TUMGray3}{rgb}{0.8, 0.8, 0.8}
\definecolor{TUMBlue4}{rgb}{0.60, 0.78, 0.92}
\definecolor{TUMIvory}{rgb}{0.85, 0.84, 0.80}
\definecolor{TUMOrange}{rgb}{0.89, 0.45, 0.13}
\definecolor{TUMGreen}{rgb}{0.64, 0.68, 0.0}
\definecolor{TUMGreenWeb}{rgb}{0.62, 0.73, 0.21}
\definecolor{TUMRedWeb}{rgb}{0.92, 0.45, 0.22}
\definecolor{NVIDIAGreen}{HTML}{76b900}
\definecolor{TUMSecondaryBlue}{HTML}{005293}
\definecolor{TUMSecondaryBlue2}{HTML}{003359}
\definecolor{TUMBlack}{HTML}{000000}
\definecolor{TUMWhite}{HTML}{FFFFFF}
\definecolor{TUMDarkGray}{HTML}{333333}
\definecolor{TUMGray}{HTML}{808080}
\definecolor{TUMLightGray}{HTML}{CCCCC6}
\definecolor{TUMAccentGray}{HTML}{DAD7CB}
\definecolor{TUMAccentOrange}{HTML}{E37222}
\definecolor{TUMAccentGreen}{HTML}{A2AD00}
\definecolor{TUMAccentLightBlue}{HTML}{98C6EA}
\definecolor{TUMAccentBlue}{HTML}{64A0C8}

\newcounter{subfigure}[figure]

\newcommand{\subfigcaptioncustom}[1]{
    \par\footnotesize(\alph{subfigure})\;
    #1
}
\newcommand{\startsubfigcustom}{\refstepcounter{subfigure}}
\newenvironment{subfigurecustom}
{%
    \refstepcounter{figure}%
    \setcounter{subfigure}{0}%
}
{%
    \addtocounter{figure}{-1}%
}
\title{\LARGE \bf
FAR-LIO: Enabling High-Speed Autonomy through \textbf{F}ast, \textbf{A}ccurate, and \textbf{R}obust LiDAR-Inertial Odometry 
}
\author{Maximilian Leitenstern$^{1*}$, Marcel Weinmann$^{1*}$, Patrick Haft$^{2}$, Tobias Lasser$^{2}$,  \\Dominik Kulmer$^{1}$, and Markus Lienkamp$^{1}$%
\thanks{$^{*}$The first two authors contributed equally to this work.}%
\thanks{$^{1}$Maximilian Leitenstern, Marcel Weinmann, Dominik Kulmer, and Markus Lienkamp are with the Institute of Automotive Technology, School of Engineering and Design, Technical University of Munich, Germany.}%
\thanks{$^{2}$Patrick Haft and Tobias Lasser are with NVIDIA Corp., USA.}%
\thanks{Corresponding authors: \small \{maxi.leitenstern,\,marcel.weinmann\}@tum.de}}%
\begin{document}
\maketitle
\copyrightnotice
\thispagestyle{empty}
\pagestyle{empty}
\begin{abstract}
Robust and accurate odometry estimation is essential in modern robotics.
In environments characterized by highly dynamic motion and sensor noise, odometry estimation becomes increasingly challenging. Autonomous racing combines both factors in an unstructured setting, where minimizing odometry latency is essential for stable closed-loop control.
This paper introduces \textit{FAR-LIO}, a highly optimized CUDA-accelerated LiDAR-inertial odometry framework developed for \textit{Fast}, \textit{Accurate}, and \textit{Robust} performance.
Our system leverages a novel CUDA-based voxel hashmap to enable parallelized nearest-neighbor search and efficient map updates.
We employ a sparsity-aware Generalized Iterative Closest Point algorithm with adaptive thresholding on top of the CUDA-based voxel hashmap with adaptive density to achieve low-latency without compromising accuracy.
An Extended Kalman Filter serves as a robust backend. It utilizes an upsampling and delay compensation strategy to fuse the LiDAR odometry with high-frequency IMU data, thereby ensuring a robust and smooth odometry output.
We evaluate \textit{FAR-LIO} across four different sensor setups, using both public datasets and data from two autonomous racecars driving at speeds of up to 250\,km/h. \textit{FAR-LIO} achieves an average \textit{6.9\%} reduction in the positional error and \textit{38.4\%} lower runtime compared to state-of-the-art baselines on target hardware using a single parameter set. This demonstrates its computational efficiency and broad applicability. To build upon our work, our code is available open-source on \href{https://github.com/TUMFTM/FAR-LIO}{https://github.com/TUMFTM/FAR-LIO}.
\end{abstract}
\section{INTRODUCTION}
\begin{figure}[!t]
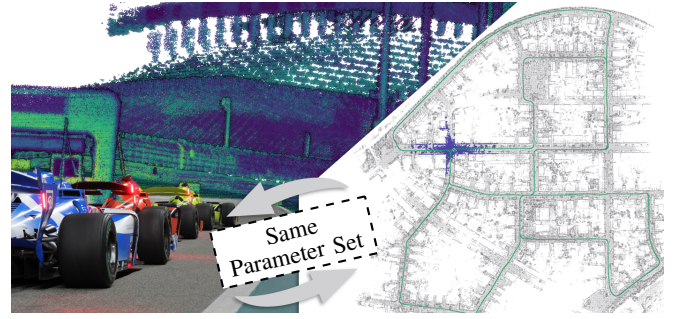

    \centering
    \include{fig/title_page}
    \vspace{-0.8cm}
    \caption{Point cloud maps of the Yas Marina Circuit (Abu Dhabi Autonomous Racing League, left) and the KITTI~00 sequence (right)~\cite{KITTI, KITTIraw}. Both were generated with \textit{FAR-LIO} using identical parameters, which demonstrates its transferability from racetrack to urban environments.}
    \vspace{-0.5cm}
    \label{fig:usp}
\end{figure}

LiDAR odometry is a core component of many mobile robotics systems, enabling safe navigation under real-time constraints. Most state-of-the-art approaches perform fine registration to align consecutive LiDAR frames and incrementally build a 3D map of their environment for localization. This process is referred to as Simultaneous Localization and Mapping (SLAM). Integrating acceleration and angular velocity measurements from an Inertial Measurement Unit (IMU) can further enhance the accuracy and robustness of these algorithms.
Many LiDAR–Inertial Odometry (LIO) frameworks, however, require extensive parameter tuning and compute resources to meet the real-time and robustness constraints necessary for real-world deployment. This is particularly challenging in domains with limited testing opportunities, extreme environmental conditions, and speeds of up to \SI{250}{km/h}, such as autonomous racing. This work demonstrates that safe, autonomous, high-speed operation requires minimizing odometry latency and strictly bounding it below critical thresholds.
Our experiments show that our highly optimized, CUDA-accelerated framework specifically addresses the limitations of many CPU-based methods, which often fail to meet real-world requirements on target hardware. Our framework integrates a CUDA-accelerated, sparsity-aware Generalized Iterative Closest Point (GICP) front-end with an Extended Kalman Filter (EKF) backend for IMU fusion. The registration utilizes a CUDA-based voxel hashmap with adaptive density, incorporating a robust kernel and the adaptive thresholding scheme from Vizzo et al. \cite{vizzo2023ral}. This way, we demonstrate \textbf{F}ast, \textbf{A}ccurate, and \textbf{R}obust performance across three datasets using four different LiDAR and IMU sensors in residential areas, on highways, and racetracks (Fig.~\ref{fig:usp}). To ensure transferability, we evaluate all algorithms on the target hardware of an autonomous racecar and using the Robot Operating System (ROS)~\cite{quigley2009, Macenski2022} to account for runtime constraints.
Our main contributions are:
\begin{itemize}
    \item We introduce and open-source \textbf{\textit{FAR-LIO}}, a highly optimized, CUDA-accelerated LiDAR-inertial odometry method with \textbf{F}ast, \textbf{A}ccurate, and \textbf{R}obust performance.
    \item We detail and ablate the core contributions of \textbf{\textit{FAR-LIO}}, including its CUDA-based voxel hashmap with adaptive density, the sparsity-aware GICP, and the EKF backend.
    \item We validate the real-world capability of our framework by demonstrating the critical role of localization latency on system robustness.
    \item We show that \textbf{\textit{FAR-LIO}} outperforms state-of-the-art methods with an average \textit{6.9\%} reduction in the positional error and \textit{38.4\%} lower runtime, using a single parameter set across public autonomous driving and high-speed autonomous racing datasets up to \SI{250}{km/h}.
\end{itemize}
\section{RELATED WORK}
Robust localization is crucial for mobile autonomy. LIO has emerged as a leading methodology, combining the geometric data from a LiDAR with the high-frequency motion updates from an IMU. In this review, we outline key LiDAR-only techniques before focusing on the more dominant LIO systems, including SLAM frameworks with loop closure. We categorize these systems by their underlying backend, either filtering or optimization, thereby deriving the research gap.
\subsection{LiDAR-Only Odometry}
LiDAR odometry algorithms employ scan registration to find the transformation between two frames, thereby incrementally deriving a trajectory. This process usually consists of two steps: data association to find correspondences and the computation of the transformation. One can distinguish between feature-based strategies, deriving the correspondences from geometric features, and direct strategies, estimating temporal correspondences from a heuristic assumption.\\%
Leveraging feature-based scan registration, \textit{LOAM}~\cite{zhang2014} can be seen as a fundamental method for LiDAR odometry and SLAM in general, building the foundation for many modern frameworks. Therefore, several approaches, such as \textit{LeGO-LOAM}~\cite{shan2018}, or \textit{F-LOAM}~\cite{wang2021} build upon it.\\%
In terms of direct scan registration, a pioneering work is the Iterative Closest Point (ICP) algorithm~\cite{besl1992}, using point-to-point distances for temporal correspondence estimation. Since its presentation in 1992, many variants have evolved to enhance its robustness and accuracy~\cite{segal2009, chen2019, koide2021, dellenbach2022, chen2022, vizzo2023ral, zheng2024, lee2024, ferrari2024}. These approaches mainly differ in how they establish correspondences (e.g., point-to-point, point-to-plane, or distribution-based), how they reject outliers, and how they optimize the transformation. The GICP~\cite{segal2009} algorithm, for instance, which is employed in this work, combines the point-to-point and point-to-plane metric within a probabilistic framework, thereby enhancing alignment robustness. Among recent advancements in robust registration, the \textit{KISS-ICP}~\cite{vizzo2023ral} introduces an adaptive thresholding method for outlier rejection, based on the deviation between the initial guess and the registered pose. Furthermore, they perform a simple scan-to-submap ICP registration using a CPU-based voxel hashmap.
\subsection{LiDAR-Interial Odometry}
LiDAR-inertial odometry frameworks often extend the scan registration techniques, as used in LiDAR-only approaches, with a backend to allow for the integration of IMU data. Existing backends may be categorized into filter-based approaches, such as Kalman Filters (KF), and optimization-based approaches, such as factor graphs~\cite{wu2024}. In the latter case, \textit{LIO-SAM}~\cite{shan2020} combines the frontend of \textit{LOAM}~\cite{zhang2014} with a factor graph optimization to directly integrate IMU pre-integration factors. \textit{LOCUS 2.0}~\cite{reinke2022}, and its predecessor~\cite{palieri2020} employ a GICP-based frontend with a sliding-window map optimization approach. A more recent work by Koide et al.~\cite{koide2024} introduces \textit{GLIM}, a full SLAM framework using a factor graph backend with a submap strategy and GPU-accelerated scan registration factors.\\%
In contrast to the computationally complex and expensive optimization approaches that often scale with map size, filter-based methods are a lightweight alternative, especially for odometry frameworks without loop closure functionality. To this end, a recent work by Wu et al.~\cite{wu2024} extends the \textit{KISS-ICP}~\cite{vizzo2023ral} by an EKF for IMU fusion, enhancing robustness. Qin et al.~\cite{qin2020} present \textit{LINS}, composing a feature-based registration frontend and an Error-State Extended Kalman Filter (ESEKF) for LiDAR-IMU fusion. In \textit{FAST-LIO}, Xu et al.~\cite{xu2021} introduce an iterated EKF for improved robustness. In their subsequent work \textit{FAST-LIO2}~\cite{xu2022}, they replace feature-based with direct scan registration, enabled by a novel incremental kd-tree \cite{cai2021}. On this foundation, Bai et al.~\cite{bai2022} present \textit{Faster-LIO}, using a data structure based on incremental voxels (\textit{iVox}), thereby showing improved runtime compared to tree structures. Instead of a KF, Chen et al.~\cite{chen2023} use a geometric observer for their approach \textit{D-LIO}, thereby guaranteeing convergence to the true state estimates. \par%
Although existing LIO systems achieve high accuracy, we found they often face two key limitations: high, unbounded computational latency that restricts use in time-critical applications, and sensitivity to parameter tuning across datasets. Moreover, few approaches explicitly address the effect of computation latency in a real-world application or demonstrate robustness in highly dynamic scenarios such as autonomous racing. \textit{FAR-LIO} closes these gaps by combining a sparsity-aware GICP, built on a CUDA-based adaptive density voxel hashmap, with an EKF that compensates for the pipeline latency. This design yields low latency without sacrificing accuracy. We open-source \textit{FAR-LIO} and validate it on both autonomous racing and public autonomous driving datasets using a single parameter set, demonstrating its generalization and robustness.
\section{METHOD}%
\begin{figure*}[!tbp]
    \centering
    \input{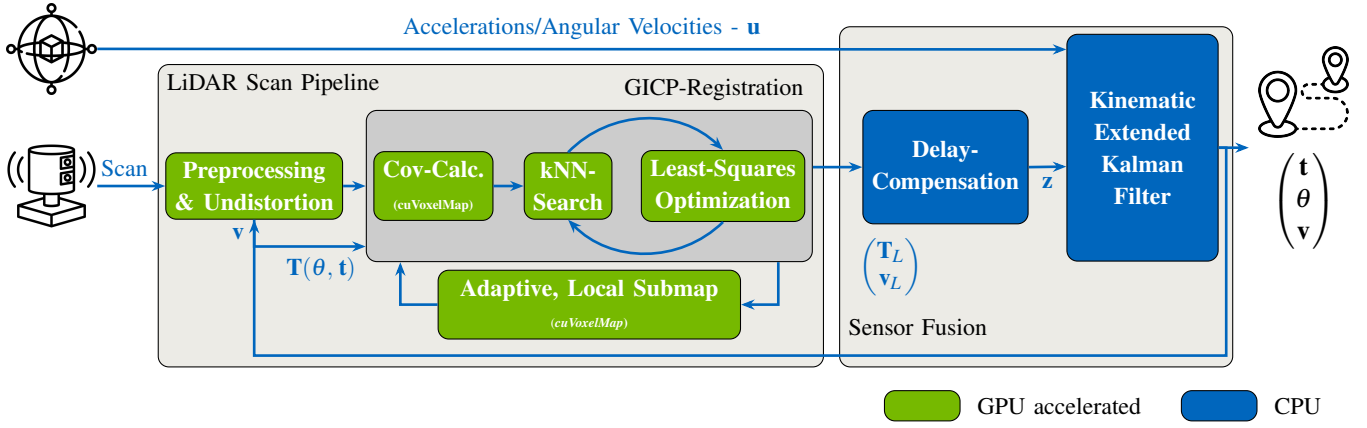}
    \vspace{0.1cm}
    \caption{Architecture Overview of \textit{FAR-LIO}. Modules colored in blue are executed on the CPU, while those colored green are GPU-accelerated using CUDA with \textit{Thrust}\protect\footnotemark[2] and \textit{cuCollections}\protect\footnotemark[1].}
    \vspace{-0.35cm}
    \label{fig:architecture}
\end{figure*}
Fig. \ref{fig:architecture} shows the architecture of \textit{FAR-LIO}. It consists of two main components that process LiDAR and IMU data, respectively. The \textit{LiDAR Scan Pipeline} operates on callback of the LiDAR scans. The registered pose is forwarded to the \textit{Sensor Fusion} component running at a fixed frequency, where it is combined with IMU data to a high-frequency odometry output $\mathbf{x}$, consisting of the pose $\mathbf{T}$ (position $\mathbf{t}$ and orientation $\mathbf{\theta}$) and the corresponding velocities $\mathbf{v}$.
\subsection{LiDAR Scan Pipeline}%
\label{subsec:scan_pipeline}%
Incoming LiDAR scans are registered to the \textit{Adaptive, Local Submap} to derive their current pose in the map frame. This pipeline extensively leverages CUDA-acceleration.
\subsubsection{Preprocessing \& Undistortion}
In the first step, the distortion of the LiDAR scan is corrected using the velocity history of the \textit{Sensor Fusion}. We perform a linear regression on both the linear and angular velocities to efficiently model the LiDAR's motion during scanning. This continuous motion model is evaluated in CUDA, effectively processing all points in parallel. Consequently, each raw point is undistorted and projected to the frame reference timestamp, which is equal to the scan's final point-timestamp.
\subsubsection{GICP-Registration}%
\begin{figure}[!htb]
    \centering
    \input{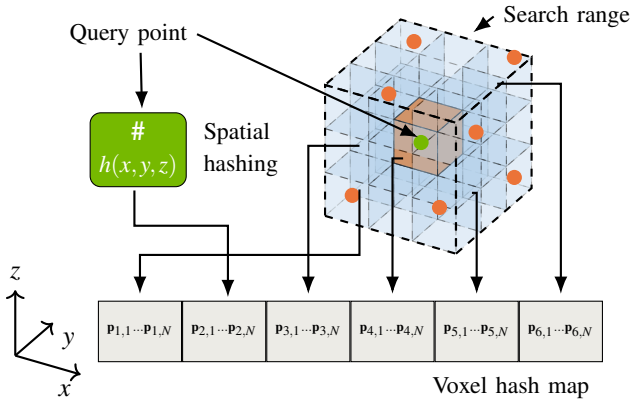}
    \vspace{-0.35cm}
    \caption{Structure of the CUDA-accelerated \textit{cuVoxelMap}.}
    \label{fig:hashmap}
\end{figure}

\label{subsubsec:gicp}%
To enable CUDA-accelerated \textit{GICP-Registration}, we introduce a novel CUDA-based voxel hashmap \textit{cuVoxelMap} based on the \textit{cuco::static\_map}\footnotemark[1]. Its design follows the \textit{iVox} paradigm, which is proven to be superior to tree-like data structures~\cite{bai2022}. We use a custom key, defined by the 3D indices of a voxel, and linear probing with a step size of \SI{1}{} to resolve hash collisions. Both the \textit{cuco::static\_map}\footnotemark[1] and our custom data structures are optimized to facilitate concurrent memory access. Employing the hash function from Niessner~et~al.~\cite{niessner2013hashing}, each key translates to a single voxel cube of size $v\,\times\,v\,\times\,v$ that may contain a fixed, maximum number $N_{max}$ of points. As shown in Fig. \ref{fig:hashmap}, the kNN-search for a given point searches the voxel containing it as well as all surrounding ones (\SI{27}{} in total).
\footnotetext[1]{\href{https://github.com/NVIDIA/cuCollections}{https://github.com/NVIDIA/cuCollections}}
\footnotetext[2]{\href{https://github.com/NVIDIA/cccl}{https://github.com/NVIDIA/cccl}}
To capture the local geometry within the covariance estimation, we use a large voxel size of $v = \SI{4}{\meter}$ in combination with a maximum of $N = 40$ regularly spaced points per voxel, resulting in a minimum spatial distance of  $\approx$\SI{65}{\centi\meter} between points. The covariance regularization leverages the Frobenius norm~\cite{koide2021}:
\begin{equation}
    \boldsymbol{\Sigma} = \left( \frac{\mathbf{C}^{-1}}{\Vert \mathbf{C}^{-1} \Vert_F} \right)^{-1} \quad \text{where} \quad \mathbf{C} = \boldsymbol{\hat{\Sigma}}~+~1e^{-3}\,\mathbf{I}.
    \label{eq:cov_reg}
\end{equation}
Here, $\hat{\boldsymbol{\Sigma}}$ is the original covariance computed from the distance to the nearest $k=10$ neighbors, while $\boldsymbol{\Sigma}$ represents the regularized covariance and $\mathbf{I}$ the identity matrix. \\
Within the \textit{GICP-Registration}, an incoming LiDAR scan $\mathcal{\mathbf{P}}^*$ is registered to the \textit{Adaptive, Local Submap} $\mathcal{\mathbf{Q}}$. The registration process starts with computing the point covariances $\boldsymbol{\Sigma}$ (\textit{Cov-Calc.}) using an instance of the \textit{cuVoxelMap}. Next, the current pose $\mathbf{T}_t$ of the \textit{Sensor Fusion} is used as the initial guess to transform the point cloud from the sensor frame into the frame of the \textit{Adaptive, Local Submap}:
\begin{equation}
    \mathcal{\mathbf{P}}^*_0 = \left\{ \mathbf{p}_{i,0} = \mathbf{T}_t\,\mathbf{p}_i \mid \mathbf{p}_i \in \mathcal{\mathbf{P}}^* \right\}.
\end{equation}
The index zero in $\mathcal{\mathbf{P}}^*_0$ indicates the initial alignment before the iterative optimization of the GICP. A single iteration $j$ starts by, for each point $\mathbf{p}_{i,j} \in \mathcal{\mathbf{P}}^*$, computing its closest Euclidean neighbor in the \textit{Adaptive, Local Submap} $\mathcal{\mathbf{Q}}$ (\textit{kNN-Search}). To reject unlikely correspondences, we use the adaptive thresholding scheme from Vizzo~et~al.~\cite{vizzo2023ral}. This yields a set of remaining correspondences $\mathcal{\mathbf{C}} = \{(\mathbf{s}_i,~\mathbf{r}_i)\}$, consisting of the source points $\mathbf{s}_i \in \mathcal{\mathbf{P}}^*$ and the corresponding reference points $\mathbf{r}_i \in \mathcal{\mathbf{Q}}$. The residual error $\mathbf{d}_i$ for a single correspondence is defined by:
\begin{equation}
    \mathbf{d}_i^{(\mathbf{T})} = \mathbf{r}_i - \mathbf{T}\mathbf{s}_i.
    \label{eq:trafo_error}
\end{equation}
The robust optimization (\textit{Least-Squares-Optimization}) then computes the transformation minimizing
\begin{align}
    \mathbf{T}_{L,j} &=\operatorname*{argmin}_\mathbf{T}\sum_{\mathbf{c}_i \in \mathcal{\mathbf{C}}} \rho\left( \mathbf{d}_i^{(\mathbf{T})^T}\left( \boldsymbol{\Sigma}_{\mathbf{r},i} + \mathbf{T}\boldsymbol{\Sigma}_{\mathbf{s},i}\,\mathbf{T}^T\right)^{-1}\mathbf{d}_i^{(\mathbf{T})}\right),
    \label{eq:optimization_target}
\end{align}
where $\rho(e)$ represents the Cauchy kernel as defined in~\cite{babin2019}:
\begin{equation}
    \rho(e) = \frac{\kappa^2}{2}~\log \left( 1\,+\,\left( \frac{e}{\kappa} \right)^2 \right).
    \label{eq:cauchy}
\end{equation}
Analogous to Vizzo~et~al.~\cite{vizzo2023ral}, the scale $\kappa$ of the robust kernel is dynamically adjusted. If a point in a sparse region of the source cloud lacks sufficient neighbors to compute a valid covariance, we utilize the special case of the GICP~\cite{segal2009}, where $\boldsymbol{\Sigma}_{\mathbf{r},i} = \mathbf{I}$ and $\boldsymbol{\Sigma}_{\mathbf{s},i} = \mathbf{0}$. This results in a sparsity-aware formulation of the GICP (SA GICP) and a point-to-point registration of these points. Lastly, the points are transformed using the optimization result $\mathbf{T}_{L,j}$:
\begin{equation}
    \mathcal{\mathbf{P}}^*_j = \left\{ \mathbf{p}_{i,j} = \mathbf{T}_{L,j}\,\,\mathbf{p}_i \mid \mathbf{p}_i \in \mathcal{\mathbf{P}}^*_{j-1} \right\}.
\end{equation}
This iterative process is repeated until either the convergence criterion is met ($\mid\mid\mathbf{T}_{L,j}\mid\mid_2~\leq~5e^{-3}$) or a maximum registration time ($t_{registration}~\geq~2\,f_L$) is reached. The time limit is defined as twice the LiDAR frequency $f_L$, while the convergence criterion is chosen empirically to ensure fast and reliable convergence under \textit{float32}-precision. Although rarely reached, the time threshold ensures that at least every second scan is processed. Even if this may lead to discarding individual frames, we found that it enhances the overall system robustness. \\%
As LiDAR-inertial systems do not feature any direct velocity measurements, we additionally compute the 3D velocity vector from the current and previous pose of the registration:
\begin{equation}
    \mathbf{v}_{L,t} = \frac{\mathbf{R}_{L,t}^{-1}\,(\mathbf{t}_t - \mathbf{t}_{t-1})}{\Delta t}.
    \label{eq:vel}
\end{equation}
$\mathbf{R}_{L,t}$ is the rotation matrix, while $\mathbf{t}_t$ and $\mathbf{t}_{t-1}$ are the translation vectors of the current and previous poses $\mathbf{T}_{L,t}$ and $\mathbf{T}_{L,t-1}$, respectively. $\Delta t$ is the time between the two poses. Finally, the odometry $\begin{bmatrix} \mathbf{T}_L~\mathbf{v}_L \end{bmatrix}^T$ is forwarded to the \textit{Sensor Fusion}.
\subsubsection{Adaptive, Local Submap}%
\label{subsubsec:submap}%
The \textit{Adaptive, Local Submap} is based on an instance of the \textit{cuVoxelMap}. It forms a local, constantly updated map around the vehicle, accumulated from the points of multiple LiDAR scans. To ensure that the submap's structure locally follows the current environment, we employ an \textit{Adaptive Submap Density} (ASMD). Before the map is updated, the average amount of points per voxel in a close range of the current scan is computed to effectively reflect the current environment. To efficiently model the decreasing point density over range, we use a linear decay (with a lower bound defined by the $k = 10$ points required to compute the covariances). In the next step, the points of the current scan $\mathcal{\mathbf{P}}^*$ are transformed to the latest pose $\mathbf{T_{L,t}}$ and integrated into the corresponding voxels of the \textit{Adaptive, Local Submap} $\mathcal{\mathbf{Q}}$, as long as the number of points inside it is below $N_{ASMD}$. After that, we recompute the covariances of the added points. Although the added points also influence the covariance of the existing points in the submap, we found this effect to be negligible, so we leave them unchanged. In the last step, voxels whose distance to the current pose $\mathbf{T_{L,t}}$ exceeds a certain limit are removed. Following Xu~et~al.~\cite{xu2022} as baseline, this limit is empirically set to \SI{1000}{\meter} to ensure reliable matching while preserving fast update times by constraining the maximum map size.
\subsection{Sensor Fusion}%
\label{subsec:sensor_fusion}%
The \textit{Sensor Fusion} module leverages an EKF to combine the registered poses of the \textit{LiDAR Scan Pipeline} with the raw IMU data. It outputs the final odometry, which is also used in the \textit{LiDAR Scan Pipeline} for undistortion and as an initial guess for the \textit{GICP-Registration}.
\subsubsection{Kinematic Extended Kalman Filter}
\label{subsubsec:ekf}%
As a robust backend, an EKF running at a fixed frequency of \SI{100}{\hertz} fuses the asynchronous IMU measurements with the registered LiDAR poses. The filter is based on the design of Goblirsch et al.~\cite{Goblirsch2024}. It uses a purely kinematic motion model, where the state consists of the position $\mathbf{t}$, orientation $\bm{\theta}$, and velocity $\mathbf{v}$ in the body frame. The measured acceleration $\mathbf{a}_m$ and angular velocity $\bm{\omega}_m$ are used directly as control inputs.
\begin{equation}
\mathbf{x} \triangleq
\begin{bmatrix}
\mathbf{t \quad \bm{\theta} \quad v}
\end{bmatrix}^T\in\mathbb{R}^{9},
\quad
\quad
\mathbf{u} \triangleq
\begin{bmatrix}
\mathbf{a}_m \quad \bm{\omega}_m
\end{bmatrix}^T\in\mathbb{R}^{6}.
\label{eq:ekf_definition}
\end{equation}
The kinematic state transition $f(\cdot)$ in Equation~\ref{eq:ekf_state_transition} employs the estimated orientation $\bm{\theta}$ through the rotation matrix $\mathbf{R}(\bm{\theta})$ and Euler angles kinematics $\mathbf{E}(\bm{\theta})$ to transform the body’s linear and angular velocities into the map frame. Furthermore, the centripetal and gravitational acceleration in the body frame $\mathbf{g}(\bm{\theta})$ is compensated:
\begin{equation}
f^{(\mathbf{x}_{k-1},\,\mathbf{u}_k)} =
\mathbf{x}_{k-1} +
\begin{bmatrix}
    \mathbf{R}^{(\bm{\theta})} \, \mathbf{v} \\[1mm]
    \mathbf{E}^{(\bm{\theta})} \, (\bm{\omega}_m - \bm{\omega}_{\text{b}}) \\[1mm]
    -\,\left[\bm{\omega}_m\right]_\times\,\mathbf{v} + (\mathbf{a}_m - \mathbf{a}_{\text{b}}) + \mathbf{g}^{(\bm{\theta})}
\end{bmatrix}.
\label{eq:ekf_state_transition}
\end{equation}
Instead of estimating the IMU bias $\mathbf{a}_{\text{b}}$ and $\bm{\omega}_{\text{b}}$ online, we perform the calibration during standstill, ensuring robust operation in high-noise environments. The predicted state is corrected using the calculated position $\mathbf{p}_L$, orientation $\bm{\theta}_L$, and velocity $\mathbf{v}_L$ obtained from LiDAR odometry, as represented in the measurement vector in Equation~\ref{eq:ekf_measurement_vector}. Following Tseng et al.~\cite{tseng2007} and Goblirsch et al.~\cite{Goblirsch2024}, we compute a set of reference angles $\bm{\theta}_{ref}$ to stabilize the estimated roll and pitch angles, thereby reducing the z-drift of our method.
\begin{equation}
\mathbf{z} \triangleq
\begin{bmatrix}
    \mathbf{t}_L \quad \bm{\theta}_L \quad \bm{\theta}_{ref} \quad \mathbf{v}_L
\end{bmatrix}^T\in\mathbb{R}^{11}.
\label{eq:ekf_measurement_vector}
\end{equation}
\subsubsection{Delay Compensation}
\label{subsubsec:delay_comp}%
Our method efficiently incorporates time-delayed measurements without introducing large jumps in the state estimate by building upon the extrapolation strategy of Larsen et al.~\cite{Larsen1998}. Equation~\ref{eq:ekf_measurement_delay_compensation} shows that delayed measurements $\mathbf{z}_s$ at timestep $s$ are projected to the current timestep $k$ by exploiting the difference between the predicted state $\mathbf{x}_{k|k-1}$ and the state history $\mathbf{x}_s$ of the EKF. Here $h(\cdot)$ denotes the nonlinear measurement model.
\begin{equation}
\mathbf{z}_k = \mathbf{z}_s + h^{(\mathbf{x}_{k|k-1})} - h^{(\mathbf{x}_s)}.
\label{eq:ekf_measurement_delay_compensation}
\end{equation}
This efficient method furthermore allows us to upsample the measurements and distribute their influence on the state correction over multiple update cycles, ensuring a smooth odometry output for subsequent control algorithms.
\section{RESULTS}
\begin{figure*}[!tb]
    \centering
    \begin{subfigurecustom}
    \begin{minipage}[t]{0.49\textwidth}
        \centering
        \startsubfigcustom
        \pgfplotsset{scaled x ticks=false}
\begin{tikzpicture}[spy using outlines={rectangle, magnification=2, connect spies}]
\node[anchor=west, minimum height=4.4cm, minimum width=0.3cm, fill=gray!10] at (3.25,2.2) {};
\node[rotate=90, anchor=west, font=\tiny] at (3.4,0) {no result};
\begin{axis}[
    name=solid,
    width=9cm,
    height=6cm,
    ybar stacked,
    bar width=0.3cm,
    bar shift=-0.25cm,
    ylabel={Runtime in \si{\milli \second}},
    ytick distance= 25,
    symbolic x coords={Fast-LIO2, D-LIO, KISS-ICP, Faster-LIO, FAR-LIO},
    xtick=data,
    x tick label style={align=center},
    xticklabels={Fast-LIO2, D-LIO, KISS-ICP, Faster-LIO, FAR-LIO\\(ours)},
    xticklabel style={rotate=30, anchor=north east, font=\small},
    legend style={at={(1,1)}, anchor=north east, legend columns=2, font=\tiny},
    ymajorgrids=true,
    yminorgrids=true,
    major grid style={gray, densely dotted},
    minor grid style={gray!30, dotted},
    enlarge x limits=0.2,
    ymin=0,
    ymax=240, %
]
\addplot+[fill=TUMBlue, draw=black] coordinates {(Fast-LIO2,35.93) (D-LIO,24.00) (KISS-ICP,0) (Faster-LIO,6.31) (FAR-LIO,3.5)}; %
\addlegendentry{Preprocessing}

\addplot+[fill=TUMBlue4, draw=black] coordinates {(Fast-LIO2,59.76) (D-LIO,44.18) (KISS-ICP,0) (Faster-LIO,37.20) (FAR-LIO,12.59)};
\addlegendentry{Registration}

\addplot+[fill=TUMGray3, draw=black] coordinates {(Fast-LIO2,24.91) (D-LIO,173.46) (KISS-ICP,0) (Faster-LIO,5.53) (FAR-LIO,3.03)}; %
\addlegendentry{Map Update}

\addplot+[fill=TUMGray, draw=black] coordinates {(Fast-LIO2,0) (D-LIO,0) (KISS-ICP,0) (Faster-LIO,0.97) (FAR-LIO,0.11)}; %
\addlegendentry{Overhead}

\coordinate (spypoint) at ($(axis cs:Faster-LIO,27.5)!.5!(axis cs:FAR-LIO,27.5)$);
\coordinate (glass) at (axis cs:Faster-LIO,125);

\end{axis}

\begin{axis}[
    at={(solid.south west)},
    anchor=south west,
    width=9cm,
    height=6cm,
    ybar stacked,
    bar width=0.3cm,
    bar shift=+0.25cm,
    symbolic x coords={Fast-LIO2, D-LIO, KISS-ICP, Faster-LIO, FAR-LIO},
    xtick=data,
    xticklabels={},
    ymin=0,
    ymax=240, %
    enlarge x limits=0.2,
    hide axis, %
]

\addplot+[fill=TUMBlue, draw=black, densely dotted] coordinates {(Fast-LIO2,14.26) (D-LIO,8.79) (KISS-ICP,3.90) (Faster-LIO,1.69) (FAR-LIO,2.98)}; %
\addplot+[fill=TUMBlue4, draw=black, densely dotted] coordinates {(Fast-LIO2,45.26) (D-LIO,12.33) (KISS-ICP,36.81) (Faster-LIO,14.54) (FAR-LIO,3.99)};
\addplot+[fill=TUMGray3, draw=black, densely dotted] coordinates {(Fast-LIO2,11.91) (D-LIO,30.07) (KISS-ICP,1.62) (Faster-LIO,1.31) (FAR-LIO,2.45)}; %
\addplot+[fill=TUMGray, draw=black, densely dotted] coordinates {(Fast-LIO2,1.01) (D-LIO,0) (KISS-ICP,2.05) (Faster-LIO,1.01) (FAR-LIO,0.17)}; %

\end{axis}

\begin{axis}[
  name=zoomaxis,
  at={(glass)},
  anchor=center,
  width=5.5cm,
  height=3.5cm,
  ybar stacked,
  bar width=0.5cm,
  bar shift=-0.45cm,
  axis background/.style={fill=white},
  ylabel={},
  ytick distance= 10,
  symbolic x coords={Faster-LIO, FAR-LIO},
  xtick=data,
  xticklabels={},
  ymajorgrids=true,
  yminorgrids=true,
  major grid style={gray, densely dotted},
  minor grid style={gray!30, dotted},
  enlarge x limits=0.5,
  ymin=0,
  ymax=55, %
]

\addplot+[fill=TUMBlue, draw=black] coordinates {(Faster-LIO,6.31) (FAR-LIO,3.5)};
\addplot+[fill=TUMBlue4, draw=black] coordinates {(Faster-LIO,37.20) (FAR-LIO,12.59)};
\addplot+[fill=TUMGray3, draw=black] coordinates {(Faster-LIO,5.53) (FAR-LIO,3.03)};
\addplot+[fill=TUMGray, draw=black] coordinates {(Faster-LIO,0.97) (FAR-LIO,0.11)};

\end{axis}

\begin{axis}[
    at={(glass)},
    anchor=center,
    width=5.5cm,
    height=3.5cm,
    ybar stacked,
    bar width=0.5cm,
    bar shift=+0.45cm,
    symbolic x coords={Faster-LIO, FAR-LIO},
    xtick=data,
    xticklabels={},
    ymin=0,
    ymax=55, %
    enlarge x limits=0.5,
    hide axis, %
]

\addplot+[fill=TUMBlue, draw=black, densely dotted] coordinates {(Faster-LIO,1.69) (FAR-LIO,2.98)};
\addplot+[fill=TUMBlue4, draw=black, densely dotted] coordinates {(Faster-LIO,14.54) (FAR-LIO,3.99)};
\addplot+[fill=TUMGray3, draw=black, densely dotted] coordinates {(Faster-LIO,1.31) (FAR-LIO,2.45)};
\addplot+[fill=TUMGray, draw=black, densely dotted] coordinates {(Faster-LIO,1.01) (FAR-LIO,0.17)};

\end{axis}

\node[draw, black, thick, minimum width=2.5cm, minimum height=1.0cm,
      anchor=center] (zoomrect) at (spypoint) {};
\draw[black, densely dashed] (zoomrect.north east) -- (zoomaxis.south east);
\draw[black, densely dashed] (zoomrect.north west) -- (zoomaxis.south west);

\end{tikzpicture}
        \vspace{-0.5cm}
        \subfigcaptioncustom{Composition of the average total computation time per frame.}
        \label{fig:runtime_composition}
    \end{minipage}
    \begin{minipage}[t]{0.49\textwidth}
        \centering
        \startsubfigcustom
        \begin{tikzpicture}
\node[anchor=west, minimum height=4.4cm, minimum width=0.3cm, fill=gray!10] at (3.25,2.2) {};
\node[rotate=90, anchor=west, font=\tiny] at (3.4,0) {no result};
\begin{axis}[
    width=9cm,
    height=6cm,
    boxplot/draw direction=y, 
    ylabel={Runtime in \si{\milli\second}},
    ytick distance= 25,
    x tick label style={align=center},
    xtick={1,2,3,4,5},
    xticklabels={Fast-LIO2, D-LIO, KISS-ICP, Faster-LIO, FAR-LIO\\(ours)},
    xticklabel style={rotate=30, anchor=north east, font=\small},
    legend style={at={(0,1)}, anchor=north west, legend columns=4, font=\tiny},
    ymajorgrids=true,
    yminorgrids=true,
    major grid style={gray, densely dotted},
    minor grid style={gray!30, dotted},
    enlarge x limits=0.1,
    ymin=0,
    ymax=240, %
]

\addplot+ [ybar, boxplot prepared={draw position=0.8,
        lower whisker=29.55, lower quartile=86.75,
        median=113.49, upper quartile=146.79,
        upper whisker=541.09,
        box extend = 0.3},
        fill=TUMIvory,
        draw=black,
        solid,
    ] coordinates {}; 
\addplot+ [boxplot prepared={draw position=1.2,
        lower whisker=4.28, lower quartile=45.67,
        median=73.80, upper quartile=92.56,
        upper whisker=807.16,
        box extend=0.3},
        fill=TUMIvory,
        draw=black,
        densely dotted,
    ] coordinates {}; 
\addplot+ [boxplot prepared={draw position=1.8,
        lower whisker=26.00, lower quartile=45.00,
        median=64.00, upper quartile=84.00,
        upper whisker=272.00,
        box extend=0.3},
        fill=TUMIvory,
        draw=black,
        solid,
    ] coordinates {}; 
\addplot+ [boxplot prepared={draw position=2.2,
        lower whisker=4.00, lower quartile=17.00,
        median=20.00, upper quartile=24.00,
        upper whisker=196.00,
        box extend=0.3},
        fill=TUMIvory,
        draw=black,
        densely dotted,
    ] coordinates {}; 
\addplot+ [boxplot prepared={draw position=3.2,
        lower whisker=0.00, lower quartile=27.00,
        median=39.00, upper quartile=56.00,
        upper whisker=1947.00,
        box extend=0.3},
        fill=TUMIvory,
        draw=black,
        densely dotted,
    ] coordinates {}; 
\addplot+ [boxplot prepared={draw position=3.8,
        lower whisker=19.00, lower quartile=40.00,
        median=47.00, upper quartile=56.00,
        upper whisker=1740.00,
        box extend=0.3},
        fill=TUMIvory,
        draw=black,
        solid,
    ] coordinates {}; 
\addplot+ [boxplot prepared={draw position=4.2,
        lower whisker=3.00, lower quartile=14.00,
        median=18.00, upper quartile=22.00,
        upper whisker=2705.00,
        box extend=0.3},
        fill=TUMIvory,
        draw=black,
        densely dotted,
    ] coordinates {}; 
\addplot+ [boxplot prepared={draw position=4.8,
        lower whisker=10.63, lower quartile=19.14,
        median=22.87, upper quartile=27.16,
        upper whisker=88.18,
        box extend=0.3},
        fill=TUMIvory,
        draw=black,
        solid,
    ] coordinates {}; 
\addplot+ [boxplot prepared={draw position=5.2,
        lower whisker=3.25, lower quartile=10.94,
        median=11.88, upper quartile=13.11,
        upper whisker=150.76,
        box extend=0.3},
        fill=TUMIvory,
        draw=black,
        densely dotted,
    ] coordinates {}; 
\end{axis}

\end{tikzpicture}
        \vspace{-0.5cm}
        \subfigcaptioncustom{Distribution of the average computation time per callback.}
        \label{fig:runtime_dist}
    \end{minipage}
    \end{subfigurecustom}
    \par\vspace*{0.1cm}
    \caption{Evaluation of computation times per LiDAR scan. Per algorithm, two results are shown: Left bars represent autonomous racing data with \SI{3}{} concatenated LiDAR scans, right bars illustrate the average of the open-source autonomous driving datasets from Table \ref{tab:datasets} with a single LiDAR.}
    \vspace{-0.35cm}
    \label{fig:runtime}
\end{figure*}
In this section, extensive experiments are presented that support our claim that \textit{FAR-LIO} outperforms state-of-the-art approaches with respect to efficiency, accuracy, and robustness.
Our approach achieves these results employing a single parameter set across various environments, including residential areas, highways, and racetracks.
We further demonstrate that the low latency of our CUDA-accelerated pipeline is critical for the approach’s robustness, thereby facilitating its deployment in real-world applications.
\subsection{Experimental Setup}
Our experiments utilize two open-source and one custom dataset, highlighted in Table \ref{tab:datasets}. These offer maximum versatility across four countries, featuring diverse environments with dynamic objects, including residential areas, highways, and racetracks, employing four different LiDAR-IMU configurations.
The first dataset combines LiDAR scans from KITTI Odometry \cite{KITTI} with low-frequency IMU data from KITTI Raw \cite{KITTIraw}, enabling an evaluation on this well-known benchmark. This experiment mostly bypasses the preprocessing stage of many algorithms by utilizing undistorted scans, allowing an analysis independent of the implemented undistortion method.
To assess the algorithms in non-European environments with a LiDAR sensor other than Velodyne, the MulRan dataset~\cite{MulRan} is employed, offering long sequences from multiple cities in South Korea. 
The final dataset was recorded by TUM Autonomous Motorsport~\cite{hoffmann2026} during the Indy Autonomous Challenge (IAC) at Autodromo Nazionale Monza (\textit{mon\_20}) and the Abu Dhabi Autonomous Racing League (A2RL) at Yas Marina Circuit (\textit{yas\_10} and \textit{yas\_20}). It features high-speed runs up to \SI{250}{km/h}, substantial IMU measurement noise, and two different LiDAR setups providing concatenated point clouds at \SI{10}{} and \SI{20}{\hertz}. \\
We compare FAR-LIO against four state-of-the-art LiDAR and LiDAR–Inertial Odometry frameworks. We select \textit{KISS-ICP}~\cite{vizzo2023ral} as a LiDAR-only baseline, as it inspired this work by offering a simple system that performs well with the same parameter set across varying conditions. Next, we evaluate \textit{FAST-LIO2}~\cite{xu2022} and \textit{Faster-LIO}~\cite{bai2022} as representatives of iterated Kalman Filter-based methods, employing the iKD-tree~\cite{cai2021} and iVox~\cite{bai2022} paradigms, respectively, with the latter also used in our approach. Finally, we include \textit{DLIO}~\cite{chen2023}, which also uses the GICP for efficient scan registration. To ensure a solid trade-off between runtime and accuracy, we evaluate \textit{KISS-ICP}~\cite{vizzo2023ral} and \textit{Faster-LIO}~\cite{bai2022} using their reported default parameters, while increasing the voxel size for \textit{FAST-LIO2}~\cite{xu2022} and \textit{DLIO}~\cite{chen2023} to \SI{1}{\meter}, allowing for a fair comparison of both runtime and accuracy across the algorithms. Despite its CUDA implementation, we exclude \textit{GLIM}~\cite{koide2024} from our evaluation. Its backend is optimized for global mapping rather than low-latency odometry, resulting in asynchronous output unsuitable for real-time applications.\\
Since our method is specifically designed for real-world deployment, all algorithms are evaluated on the \textit{dSpace AUTERA Autobox} with an \textit{Intel\textregistered~Xeon\textregistered~D-2166NT CPU} (\SI{12}{}~×~\SI{3}{\giga\hertz}) and an \textit{NVIDIA RTX A5000}, the official compute platform of the IAC. The algorithms are executed on four CPU cores with a fixed frequency of \SI{3}{\giga\hertz}, representing the maximum compute that can be allocated to the localization task in a real-world deployment on this system, or on the GPU. Furthermore, all experiments are conducted in a non-deterministic simulation setup using ROS~\cite{quigley2009, Macenski2022}. To account for variance in the results, each experiment is repeated three times, and the best performance is reported.
\begin{table}[!htb]
\caption{Details of all datasets used for the evaluation.}
\resizebox{\columnwidth}{!}{%
\begin{tabular}{@{}lccll@{}}
\toprule \toprule
Dataset & Seq. & \begin{tabular}[c]{@{}l@{}}Length\\ {(}\si{\kilo \meter}{)}\end{tabular} & LiDAR (\si{\hertz}) & IMU (\si{\hertz}) \\
\midrule
\begin{tabular}[c]{@{}l@{}}Autonomous \\ Racing \end{tabular} & \begin{tabular}[c]{@{}l@{}}\small{yas\_10} \\ \small{yas\_20} \\ \small{mon\_20} \end{tabular} & \begin{tabular}[c]{@{}c@{}}8.3 \\ 10.5 \\ 10.0 \end{tabular} &
\begin{tabular}[c]{@{}l@{}}3x Seyond Falcon (\si{10}{}) \\ 3x Seyond Falcon (\si{20}{}) \\ 3x Luminar Iris (20)\end{tabular} &
\begin{tabular}[c]{@{}l@{}}Vectornav VN-310 (\si{800}{}) \\Vectornav VN-310 (\si{800}{}) \\ Epson G370N \end{tabular} \\
\midrule
\begin{tabular}[c]{@{}l@{}}KITTI \cite{KITTI, KITTIraw}\end{tabular} & \begin{tabular}[c]{@{}c@{}}00 \\ to 10\end{tabular} & \begin{tabular}[c]{@{}c@{}}0.4\\to 5.1\end{tabular} & Velodyne HDL-64 (\si{10}{}) & OXTS RT 3003 (\si{10}{}) \\
\midrule
\begin{tabular}[c]{@{}l@{}}MulRan\cite{MulRan}\end{tabular} & all & \begin{tabular}[c]{@{}c@{}}4.9\\to 23.4\end{tabular} & Ouster OS1-64 (\si{10}{}) & Xsens MTi-300 (\si{100}{}) \\
\bottomrule
\end{tabular}%
}
\label{tab:datasets}
\end{table}
\vspace{-0.35cm}
\subsection{Runtime Experiments}
\begin{table*}[!htb]
\centering
\begin{threeparttable}
\caption{Root-mean-squared-error (RMSE) of the average positional error (APE) in \si{\meter} and the relative positional error (RPE) in \si{\percent} per \SI{100}{\meter} for evaluated datasets in a \textit{ROS}\cite{quigley2009, Macenski2022} simulation environment computed with \textit{evo}~\cite{grupp2017evo}. Bold values denote the best result for a sequence, while "x" indicates no valid result.}
\begin{tabular}{@{}llllllllllll@{}}
\toprule \toprule
\multirow{2}{*}{Dataset} & \multirow{2}{*}{Sequence} & \multicolumn{2}{c}{\begin{tabular}[c]{@{}l@{}}Fast-LIO2~\cite{xu2022}\end{tabular}} & \multicolumn{2}{c}{\begin{tabular}[c]{@{}l@{}}D-LIO~\cite{chen2023}\end{tabular}} & \multicolumn{2}{c}{\begin{tabular}[c]{@{}l@{}}KISS-ICP~\cite{vizzo2023ral}\end{tabular}} & \multicolumn{2}{c}{\begin{tabular}[c]{@{}l@{}}Faster-LIO~\cite{bai2022}\end{tabular}} & \multicolumn{2}{c}{\begin{tabular}[c]{@{}l@{}}FAR-LIO (Ours)\end{tabular}} \\
\cmidrule(l){3-12} & & APE & RPE & APE & RPE & APE & RPE & APE & RPE & APE & RPE \\
\midrule
\multirow{3}{*}{Autonomous Racing~~} & $yas\_10$~~~~~~~~ & 37.24 & 2.14 & 19.70 & 4.34 & x & x  & 36.00 & 3.30 & \textbf{15.68} & \textbf{1.67} \\
                     & $yas\_20$ & 50.13 & 2.08 & 45.23 & 3.31 & x & x & 55.71 & 3.07 & \textbf{25.62} & \textbf{1.84} \\
                     & $mon\_20$ & \textbf{262.16} & 4.44 & x & x & x & x & 475.28 & 7.06 & 297.39 & \textbf{2.74} \\
\midrule
\multirow{10}{*}{KITTI~\cite{KITTI,KITTIraw}} & $kitti\_00$ & 10.96 & \textbf{1.04} & 18.43 & 1.09 & 14.47 & 1.14 & 13.23 & 1.41 & \textbf{9.79} & 1.31 \\
                        & $kitti\_01$ & x & x & 118.81 & 1.25 & 122.88 & 1.13 & x & x & \textbf{108.25} & \textbf{1.09} \\
                        & $kitti\_02$ & 39.30 & \textbf{1.35} & 37.48 & 1.38 & 37.00 & 1.36 & 44.42 & 2.26 & \textbf{32.9} & \textbf{1.35} \\
                        & $kitti\_04$ & 3.96 & 1.36 & 3.77 & 1.37 & 4.09 & 1.22 & 9.03 & 2.45 & \textbf{3.60} & \textbf{1.21} \\
                        & $kitti\_05$ & 6.33 & 1.13 & 4.53 & 1.14 & 6.02 & 1.08 & 6.15 & 1.60 & \textbf{3.45} & \textbf{0.94} \\
                        & $kitti\_06$ & \textbf{3.06} & 0.94 & 3.54 & 0.90 & 4.21 & 0.99 & 3.34 & 0.92 & 3.57 & \textbf{0.89} \\
                        & $kitti\_07$ & 4.73 & 0.98 & 3.08 & 0.81 & \textbf{0.53} & 0.86 & x & x & 1.37 & \textbf{0.80} \\
                        & $kitti\_08$ & 20.36 & 1.58 & 18.64 & 1.59 & 24.92 & \textbf{1.51} & 20.33 & 2.60 & \textbf{16.36} & \textbf{1.51} \\
                        & $kitti\_09$ & 16.46 & 1.09 & 11.69 & 1.04 & 148.49 & 8.21 & 121.69 & 2.50 & \textbf{8.05} & \textbf{0.89} \\
                        & $kitti\_10$ & x & x & 13.90 & 1.45 & 17.60 & 1.46 & x & x & \textbf{4.38} & \textbf{1.36} \\
\midrule
\multirow{4}{*}{MulRan~\cite{MulRan}} & $DCC$ & 28.73 & 3.23 & 18.93 & 3.64 & 31.86 & 3.04 & 68.98 & 6.61 & \textbf{18.23} & \textbf{3.02} \\
                      & $KAIST$ & 40.23 & 2.88 & 30.24 & 3.41 & 35.91 & \textbf{2.65} & 81.42 & 12.25 & \textbf{28.11} & 2.71 \\
                      & $Riverside$ & 99.30 & 3.21 & 85.19 & 3.21 & 87.01 & 2.92 & 336.42 & 13.83 & \textbf{65.95} & \textbf{2.91} \\
                      & $Sejong$ & 2543.67 & 3.82 & x & x & 2660.90 & \textbf{3.57} & 4507.82 & 36.28 & \textbf{2280.18} & 6.54 \\
\bottomrule
\end{tabular}%
\label{tab:benchmark}
\end{threeparttable}
\vspace{-0.35cm}
\end{table*}
Fig. \ref{fig:runtime} presents the computation times of the LiDAR scan pipelines for the selected algorithms. We distinguish the runtimes of the autonomous racing dataset, which employs three LiDARs producing approximately \SI{200000}{} points per concatenated scan, from those of open-source datasets, which use a single LiDAR with fewer than \SI{100000}{} points per scan.
In Fig. \ref{fig:runtime_composition}, we analyze the composition of the average runtime per frame, divided into preprocessing (typically including motion deskewing and downsampling), frame-to-submap registration, local submap update, and additional overhead, which together determine the total computation time.
With a total callback time of \SI{19.23}{\milli \second} on autonomous racing data and \SI{9.59}{\milli \second} on the open-source datasets, \textit{FAR-LIO} achieves the fastest computation times, outperforming the next-best method by an average of \textit{38.4\%}. \textit{FAR-LIO} demands \SI{2}{\giga \byte} VRAM for preallocation, utilizing less than \SI{10}{\percent} of the GPU on open-source benchmarks and ~\SI{20}{\percent} on the autonomous racing dataset. Although the average runtime increases by roughly \SI{2}{} on the autonomous racing dataset due to more registered points, our local submap update time remains nearly constant. \textit{FAR-LIO} consistently operates well below the \SIrange{10}{20}{\hertz} sensor frequency across all datasets.
Similar to our method, \textit{Faster-LIO} demonstrates low runtimes ranging from \SI{50.01}{\milli \second} on the autonomous racing dataset to \SI{18.55}{\milli \second} on the open-source datasets. In contrast, \textit{Fast-LIO2}, \textit{D-LIO}, and \textit{KISS-ICP} are unable to operate within the sensor frequency or diverge on the autonomous racing dataset. This highlights the limited robustness and scalability of CPU-based algorithms on high-density LiDAR frames using a single parameter set. \\%
To further analyze the computational efficiency of the algorithms, we evaluate the distributions of the total callback time in Fig. \ref{fig:runtime_dist}. For \textit{D-LIO}, the map update is excluded from this analysis, as it is executed in a separate thread.
Across all datasets, \textit{FAR-LIOs} computation times are tightly clustered around the median, with the maximum values primarily reflecting peaks caused by submap memory allocations or not converged frames. This is achieved through the \textit{cuVoxelMap} with adaptive density, which scales efficiently with the number of points, providing the EKF backend with frequent, low-latency updates for high robustness.
The CPU-based algorithms exhibit wider computation time distributions, especially for the dense point clouds in the autonomous racing dataset. While \textit{Faster-LIO} has a higher median, its distribution is the only one comparable to that of our method, highlighting the effectiveness of a voxel-based data structure in achieving consistent lookup and update times.
\subsection{Accuracy Experiments}
Although computational efficiency is the core focus of our approach, we do not sacrifice accuracy. Therefore, we evaluate its absolute positional error (APE) and relative positional error (RPE) against the selected state-of-the-art algorithms in Table \ref{tab:benchmark}. While the APE indicates the global deviation of an LIO system, the RPE provides information about local consistency. It can be seen that \textit{FAR-LIO} outperforms all other approaches across the datasets, ranking first or second in the majority of sequences. \\%
The autonomous racing dataset features high accelerations and turn rates, leading to strong scan distortions. Since \textit{KISS-ICP} relies solely on a constant-velocity model for the initial guess and does not incorporate IMU data, it diverged across all sequences.
\textit{FAR-LIO} shows the lowest RPE across all sequences, demonstrating superior local consistency, and the lowest APE for both \textit{yas\_10} and \textit{yas\_20}. \textit{D-LIO} follows closely on the \textit{yas} sequences in terms of the APE, but fails to produce a valid result on \textit{mon\_20}. Both \textit{Fast-LIO2} and \textit{Faster-LIO} successfully complete all sequences, with \textit{Fast-LIO2} achieving the lowest APE on \textit{mon\_20}.\\%
For the KITTI dataset~\cite{KITTI, KITTIraw}, recorded in Karlsruhe, Germany, we evaluate the sequences $00$ to $10$, where the ground truth is publicly available as part of the KITTI Odometry Benchmark. While \textit{D-LIO}, \textit{KISS-ICP}, and \textit{FAR-LIO} produce valid and comparable results among all sequences, it is noticeable that both \textit{Fast-LIO2} and \textit{Faster-LIO} fail to complete sequences $01$ and $10$. Following their similar design, this is likely caused by the low-frequency IMU data and initial velocity of the sequence. Nevertheless, it can be seen that \textit{FAR-LIO} shows the best results for the majority of sequences, resulting in the best average performance on this dataset.
Lastly, we evaluate on the MulRan dataset~\cite{MulRan} recorded in South Korea, featuring \SI{4}{} sequences in different environments, including the particularly long \textit{Sejong} sequence. Following the trend of the autonomous racing and KITTI~\cite{KITTI, KITTIraw} dataset, \textit{FAR-LIO} yields the best overall accuracy, recording the lowest APE on all sequences and the lowest RPE on \textit{DCC} and \textit{Riverside}. \textit{Fast-LIO2}, \textit{KISS-ICP}, and \textit{D-LIO} perform comparably. \textit{D-LIO} yields robust APEs but fails on \textit{Sejong}, whereas \textit{KISS-ICP} achieves the lowest RPE on \textit{KAIST} and \textit{Sejong}. \textit{Faster-LIO} shows degraded results for this dataset. The relatively large RPE of \textit{FAR-LIO} on the \textit{Sejong} sequence results from pitch misalignment between the estimated reference angles and the local submap, which occasionally leads to inaccurate registration. It is noticeable that the \textit{KISS-ICP} is on par with the best results for both the KITTI and MulRan, indicating the great importance of the scan registration for the final odometry in this case. \\%
In summary, \textit{FAR-LIO} outperforms other state-of-the-art methods, reducing the APE and RPE by a combined average of \textit{6.9\%} across the datasets using a uniform parameter set and without model-specific LiDAR handlers. Specifically, when compared to \textit{Faster-LIO}, the only method with comparable computation times, \textit{FAR-LIO} achieves substantially better results. This demonstrates that its emphasis on efficiency does not sacrifice accuracy. Unlike the other approaches, which deliver an invalid result for at least one sequence, our method completes all sequences, highlighting its robustness and suitability for real-world deployment.
\subsection{Ablation Studies}
To validate the proposed architecture and quantify the impact of LiDAR odometry latency on system accuracy and robustness, we perform component-wise ablation studies and evaluate performance under varying simulated delays.
\subsubsection{Key Design Choices}
\begin{table}[!h]
\centering
\caption{Component-wise ablation study on the Yas Marina Autonomous Racing dataset, based on the RMSE of the APE in \si{\meter} and the RPE in \si{\percent} per \SI{100}{\meter} }
\begin{tabular}{@{}ccccc|llll@{}}
\toprule \toprule
\multirow{2}{*}{EKF} & \multirow{2}{*}{DC} & \multirow{2}{*}{MU} & \multirow{2}{*}{\begin{tabular}{@{}c@{}}SA \\ GICP\end{tabular}} & \multirow{2}{*}{ASMD} & \multicolumn{2}{c}{\begin{tabular}[c]{@{}l@{}}$yas\_10$\end{tabular}} & \multicolumn{2}{c}{\begin{tabular}[c]{@{}l@{}}$yas\_20$\end{tabular}} \\
\cmidrule(l){6-9} &&&&& APE & RPE & APE & RPE \\
\midrule
&&&&& 66.32 & 6.94 & 192.7 & 26.83 \\
\checkmark &&&&& x & x & 45.34 & 1.94\\
\checkmark & \checkmark &&&& 21.46 & 1.97 & 46.17 & 1.94  \\
\checkmark & \checkmark & \checkmark &&& 28.09 & \textbf{1.66} & 48.18 & 1.87 \\
\checkmark & \checkmark & \checkmark & \checkmark && 16.11 & \textbf{1.66} & 37.47 & \textbf{1.81} \\
\checkmark & \checkmark & \checkmark & \checkmark & \checkmark & \textbf{15.68} & 1.67 & \textbf{25.62} & 1.84 \\
\bottomrule
\end{tabular}%
\label{tab:ablation}
\begin{tablenotes}
\item[1] DC = Delay Compensation, MU = Motion Undistortion,
\item[2] SA GICP = Sparsity-Aware GICP, ASMD = Adaptive Submap Density
\end{tablenotes}
\end{table}
We evaluate the influence of FAR-LIO's key components in~Table~\ref{tab:ablation} using the \textit{yas} sequences of the Autonomous Racing dataset. These present a challenging combination of substantial IMU noise, severe LiDAR distortion from velocities exceeding \SI{250}{km/h}, and intense computational load. Table~\ref{tab:ablation} shows that coupling the EKF-backend with delay compensation yields a more robust baseline than the constant-velocity assumption utilized by KISS-ICP~\cite{vizzo2023ral}. The ablation study further demonstrates that the motion undistortion module effectively reduces the RPE, while the sparsity-aware GICP and adaptive submap density significantly improve the APE, mitigating long-term drift.
\subsubsection{Importance of Odometry Latency}
\begin{figure}[!tbp]
    \centering
    \tikzmath{
  function symlog(\x,\a){
    \yLarge = ((\x>\a) - (\x<-\a)) * (ln(max(abs(\x/\a),1)) + 1);
    \ySmall = (\x >= -\a) * (\x <= \a) * \x / \a ;
    return \yLarge + \ySmall ;
  };
  function symexp(\y,\a){
    \xLarge = ((\y>1) - (\y<-1)) * \a * exp(abs(\y) - 1) ;
    \xSmall = (\y>=-1) * (\y<=1) * \a * \y ;
    return \xLarge + \xSmall ;
  };
}

\begin{tikzpicture}
  \def\basis{100} %
  \begin{axis}[
    xlabel={Average simulated odometry delay in \si{\milli \second}},
    ylabel={APE (RMSE) increase in \si{\percent}},
    xmin=0, xmax=120,
    y coord trafo/.code={\pgfmathparse{symlog(#1,\basis)}\pgfmathresult},
    y coord inv trafo/.code={\pgfmathparse{symexp(#1,\basis)}\pgfmathresult},
    unbounded coords=discard,
    filter discard warning=false,
    ytick = {0,100,1000,10000},
    yticklabels={$0$, $10^{2}$, $10^{3}$, $10^{4}$},
    grid=major,
    width=\linewidth,
    height=5.5cm,
    legend pos=north west,
    legend style={
        font=\scriptsize,
        draw=none,
        align=left
    },
    legend cell align=left
  ]

\addplot [
    TUMBlue,
    mark=*,
    mark size=1pt,
    line width=0.2pt,
    smooth
] table [x=delay_odometry_ma, y=rmse_ma, col sep=comma] {fig/results_fixed/results_mon_20250629_0_slam.sc_ma3.csv};
\addplot [
    TUMBlue,
    mark=*,
    mark size=1pt,
    line width=0.2pt,
    smooth,
    forget plot
] table [x=delay_odometry_ma, y=rmse_ma, col sep=comma] {fig/results_fixed/results_yas_20250502_0_slam.sc_ma3.csv};
\addplot [
    TUMBlue,
    mark=*,
    mark size=1pt,
    line width=0.2pt,
    smooth,
    forget plot
] table [x=delay_odometry_ma, y=rmse_ma, col sep=comma] {fig/results_fixed/results_yas_20250504_0_slam_raw.sc_ma3.csv};

\addplot [
    TUMGray,
    mark=*,
    mark size=1pt,
    line width=0.2pt,
    smooth
] table [x=delay_odometry_ma, y=rmse_ma, col sep=comma] {fig/results_fixed/results_mulran_DCC01.sc_ma3.csv};
\addplot [
    TUMGray,
    mark=*,
    mark size=1pt,
    line width=0.2pt,
    smooth,
    forget plot
] table [x=delay_odometry_ma, y=rmse_ma, col sep=comma] {fig/results_fixed/results_mulran_DCC02.sc_ma3.csv};
\addplot [
    TUMGray,
    mark=*,
    mark size=1pt,
    line width=0.2pt,
    smooth,
    forget plot
] table [x=delay_odometry_ma, y=rmse_ma, col sep=comma] {fig/results_fixed/results_mulran_DCC03.sc_ma3.csv};
\addplot [
    TUMBlue4,
    mark=*,
    mark size=1pt,
    line width=0.2pt,
    smooth,
] table [x=delay_odometry_ma, y=rmse_ma, col sep=comma] {fig/results_fixed/results_mulran_KAIST01.sc_ma3.csv};
\addplot [
    TUMBlue4,
    mark=*,
    mark size=1pt,
    line width=0.2pt,
    smooth,
    forget plot
] table [x=delay_odometry_ma, y=rmse_ma, col sep=comma] {fig/results_fixed/results_mulran_KAIST02.sc_ma3.csv};
\addplot [
    TUMBlue4,
    mark=*,
    mark size=1pt,
    line width=0.2pt,
    smooth,
    forget plot
] table [x=delay_odometry_ma, y=rmse_ma, col sep=comma] {fig/results_fixed/results_mulran_KAIST03.sc_ma3.csv};

\legend{Autonomous Racing, MulRan (DCC), MulRan (KAIST)}
\draw[TUMGray, dashed, line width=1pt] (axis cs:13.04,-100.0) -- (axis cs:13.04,10000);
\draw[TUMBlue4, dashed, line width=1pt] (axis cs:12.74,-100.0) -- (axis cs:12.74,10000);
\draw[TUMBlue, dashed, line width=1pt] (axis cs:24.07,-100.0) -- (axis cs:24.07,10000);
\draw[-Latex, thick,line width=0.8pt] (axis cs:30,2500) node[right, align=left]
{} -- (axis cs:24.07,1000);
\draw[-Latex, thick,line width=0.6pt] (axis cs:30,2500) node[right, align=left] {}
 -- (axis cs:12.74,1000);
\end{axis}
\node[rectangle, align=left] at (2.8cm, 1.7cm) {\scriptsize{Real average} \\ \scriptsize{delay of \textit{FAR-LIO}}};
\end{tikzpicture}
    \vspace{-0.5cm}
    \caption{Increase of the RMSE of the average positional error (APE) caused by simulated additional odometry delay. Each data point represents a single simulation with an additional delay added to the registered pose.}
    \label{fig:latency}
\end{figure}
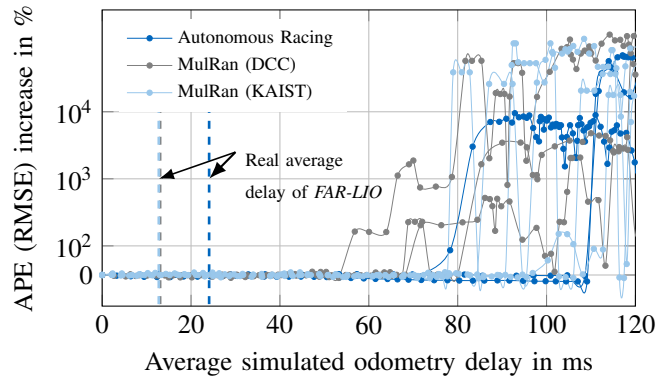

Minimizing odometry latency has proven critical for robust localization and, consequently, closed-loop vehicle control when deploying \textit{FAR-LIO} in autonomous racing. These findings are supported by the results shown in~Fig.~\ref{fig:latency}. By systematically scaling the LiDAR scan pipeline latency in simulation using RSLCPP \cite{sagmeister2026}, we find that the APE remains stable up to a sequence-dependent threshold of \SIrange{60}{100}{\milli\second}, beyond which it rises sharply, indicating divergence of the odometry estimation. This effect arises because the worst-case runtime can far exceed the average, leading to large delays to be compensated, discarded LiDAR frames, and ultimately an unstable odometry estimate. As shown in ~Fig.~\ref{fig:runtime_dist}, our runtime results demonstrate that the actual execution time of \textit{FAR-LIO} remains well within a safe region, with a substantial margin to the previously identified critical thresholds. This underlines the robustness of our approach in providing real-time, high-frequency odometry suitable for fast and stable control.
\section{CONCLUSION}
This work presents \textit{FAR-LIO}, a \textbf{F}ast, \textbf{A}ccurate, and \textbf{R}obust LiDAR-inertial odometry framework. The proposed method combines a CUDA-accelerated, sparsity-aware GICP, built on a CUDA-based voxel hashmap featuring adaptive density, with an EKF backend, enabling efficient and robust odometry estimation. Utilizing a single parameter set, \textit{FAR-LIO} outperforms state-of-the-art baselines across multiple open-source autonomous driving and real-world autonomous racing datasets, yielding an average \textit{6.9\%} reduction in the positional error and a \textit{38.4\%} decrease in runtime. We evaluated \textit{FAR-LIO} alongside several state-of-the-art algorithms on the compute hardware of an autonomous racecar, demonstrating its capability to handle diverse sensor configurations and environments, including residential areas, highways, and racetracks. We further employ simulation results to demonstrate the robustness of our method, supporting the claim that low latency is critical for robust odometry estimation. By open-sourcing our framework, which has been rigorously validated through real-world deployment on an autonomous racecar, we aim to provide a robust and simple baseline for future research in LiDAR-Inertial odometry.
\section*{ACKNOWLEDGMENT}
This work was supported by basic research funds of the Institute of Automotive Technology. The authors gratefully acknowledge the support of the Abu Dhabi Autonomous Racing League and the Indy Autonomous Challenge for data recording and evaluation. We further thank Daniel Jünger from NVIDIA Corp., for assistance during the development of the \textit{cuVoxelMap}. AI tools (GPT5.2, Gemini 3 Pro) were used exclusively for editorial, code, and documentation refinement of author-provided input.
\bibliographystyle{IEEEtran}
\bibliography{references}
\end{document}